\documentclass[conference]{IEEEtran}
\usepackage{subcaption}
\usepackage{cite}
\usepackage{amsmath,amssymb,amsfonts}
\usepackage{algorithmic}
\usepackage{graphicx}
\usepackage{textcomp}
\usepackage{xcolor}
\usepackage{float}
\usepackage{changes}
\usepackage{multirow}

\IEEEoverridecommandlockouts

\def\BibTeX{{\rm B\kern-.05em{\sc i\kern-.025em b}\kern-.08em
    T\kern-.1667em\lower.7ex\hbox{E}\kern-.125emX}}
\begin{document}

\title{
\textsc{\textbf{DFCon:} Attention-Driven Supervised Contrastive Learning for Robust Deepfake Detection} \\
{\large \textbf{IEEE SPS Signal Processing Cup: Team Straw Hats}} 
}

\author{
\IEEEauthorblockN{
MD Sadik Hossain Shanto\textsuperscript{*}, 
Mahir Labib Dihan\textsuperscript{*}, 
Souvik Ghosh\textsuperscript{*}, 
Riad Ahmed Anonto\textsuperscript{*}, \\
Hafijul Hoque Chowdhury\textsuperscript{*}, 
Abir Muhtasim\textsuperscript{*}, 
Rakib Ahsan\textsuperscript{*}, 
MD Tanvir Hassan\textsuperscript{*}, 
MD Roqunuzzaman Sojib\textsuperscript{*}, \\
Sheikh Azizul Hakim\textsuperscript{\dag}, 
M. Saifur Rahman\textsuperscript{\ddag} 
}
\IEEEauthorblockA{
\\
\textit{Department of Computer Science and Engineering}, \\
\textit{Bangladesh University of Engineering and Technology}, Dhaka, Bangladesh \\
\\
Emails: \{shantosadikrglhs@gmail.com, mahirlabibdihan@gmail.com, souvik7701@gmail.com, riadahmedanonto355@gmail.com, \\
nabidhasan987@gmail.com, auntor505@gmail.com, iamrakib242@gmail.com, saad7557.7557a@gmail.com, \\
sojibxaman439@gmail.com, hakim@cse.buet.ac.bd, mrahman@cse.buet.ac.bd\}
}
\thanks{\rule{0.5\columnwidth}{0.3pt}}
\thanks{\textsuperscript{*}Equal Contribution}
\thanks{\textsuperscript{\dag}Graduate Mentor}
\thanks{\textsuperscript{\ddag}Supervisor}
}
\maketitle

\begin{abstract}
This report presents our approach for the IEEE SP Cup 2025: Deepfake Face Detection in the Wild (DFWild-Cup), focusing on detecting deepfakes across diverse datasets. Our methodology employs advanced backbone models, including MaxViT, CoAtNet, and EVA-02, fine-tuned using supervised contrastive loss to enhance feature separation. These models were specifically chosen for their complementary strengths. Integration of convolution layers and strided attention in MaxViT is well-suited for detecting local features. In contrast, hybrid use of convolution and attention mechanisms in CoAtNet effectively captures multi-scale features. Robust pretraining with masked image modeling of EVA-02 excels at capturing global features. After training, we freeze the parameters of these models and train the classification heads. Finally, a majority voting ensemble is employed to combine the predictions from these models, improving robustness and generalization to unseen scenarios. The proposed system addresses the challenges of detecting deepfakes in real-world conditions and achieves a commendable accuracy of 95.83\% on the validation dataset.
\end{abstract}


\section{Introduction}

The proliferation of deepfake technology has raised significant concerns regarding its misuse for spreading misinformation and undermining security. Detecting deepfakes is a critical challenge, and the IEEE SP Cup 2025 provides a platform to address this pressing issue. In this report, we present a robust framework for deepfake detection by ensembling three advanced pretrained backbone models: MaxViT\cite{tu2022maxvitmultiaxisvisiontransformer}, CoAtNet\cite{dai2021coatnetmarryingconvolutionattention}, and EVA-02\cite{Fang_2024}. MaxViT is a hybrid model that integrates convolutional layers with Vision Transformers\cite{dosovitskiy2021imageworth16x16words}, employing multi-axis self-attention (Max-SA) to capture both local and global spatial features efficiently. It reduces the complexity of self-attention while maintaining non-locality, offering a balance between computational efficiency and performance. These properties make MaxViT particularly well-suited for detecting fine-grained manipulations in deepfake images.

CoAtNet\cite{dai2021coatnetmarryingconvolutionattention} leverages two key insights: unifying depthwise convolution and self-attention through simple relative attention and vertically stacking convolution and attention layers to improve generalization, capacity, and efficiency. EVA-02\cite{Fang_2024}, with its masked image modeling and vision-language alignment, provides a global understanding of image content. Together, these models form a balanced ensemble that effectively captures diverse features, enhancing overall accuracy across various datasets. We have employed offline and online data augmentation techniques to improve performance further. Offline augmentation generates diverse variations of images before training, while online augmentation applies transformations dynamically during training. These strategies enhance model generalization across diverse datasets and prevent overfitting.

\begin{figure*}[h!]
    \centering
    \begin{subfigure}[t]{0.238\textwidth}
        \centering
        \includegraphics[width=\textwidth]{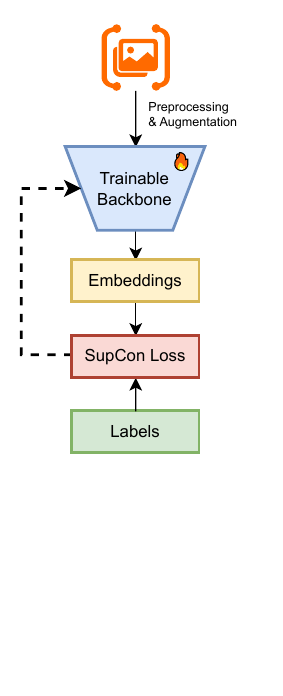}
        \caption{Stage 1: Trainable Backbone optimized with supervised contrastive loss.}
        \label{fig:stage1}
    \end{subfigure}
    \hfill
    \begin{subfigure}[t]{0.238\textwidth}
        \centering
        \includegraphics[width=\textwidth]{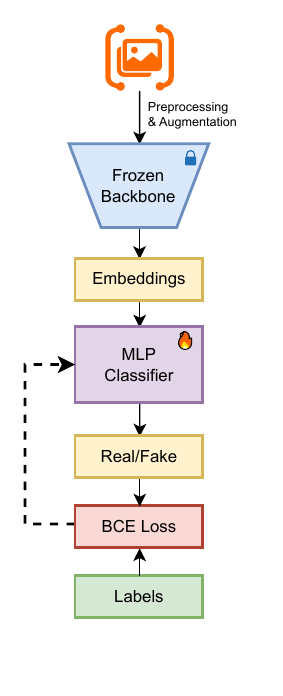}
        \caption{Stage 2: Frozen Backbone with BCE loss.}
        \label{fig:stage2}
    \end{subfigure}
    \hfill
    \begin{subfigure}[t]{0.467\textwidth}
        \centering
        \includegraphics[width=\textwidth]{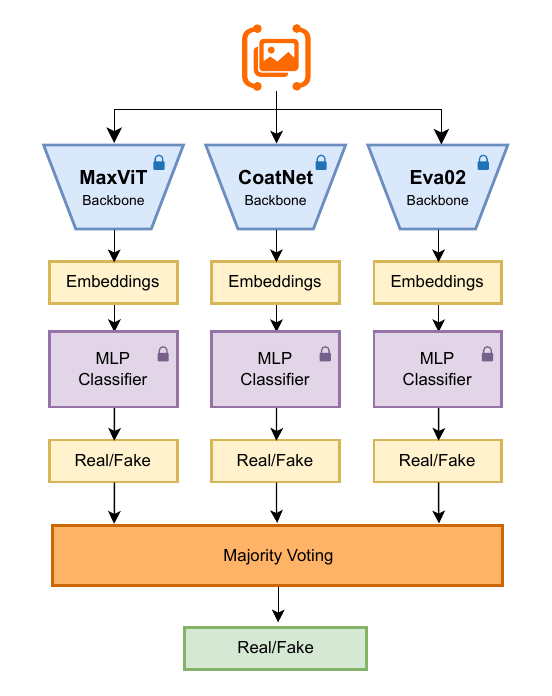}
        \caption{Stage 3: Ensemble of MaxViT, CoAtNet, and EVA-02 with majority voting.}
        \label{fig:stage3}
    \end{subfigure}
    \caption{Overview of the proposed framework across three stages.}
    \label{fig:proposed_framework}
\end{figure*}

Supervised contrastive loss\cite{khosla2021supervisedcontrastivelearning} is used to ensure distinct separation between real and fake images in the embedding space. By bringing embeddings of similar samples closer and pushing apart those of dissimilar ones, the loss function improves the quality of the feature representations, providing a strong foundation for accurate classification. Finally, the predictions from MaxViT, CoAtNet, and EVA-02 are combined through an ensemble strategy. This approach leverages the models' complementary strengths for robust and accurate deepfake detection across diverse competition scenarios.

\section{Methodology}

\subsection{Competition Dataset}
The DFWild-Cup 2025 dataset consists of images from DeepfakeBench\cite{DeepfakeBench_YAN_NEURIPS2023} evaluation framework, providing a diverse collection of real and fake facial images to assess the generalization capabilities of deepfake detection systems. The training set includes 262,160 images, consisting of 42,690 real and 219,470 fake samples, and the validation set comprises evenly distributed 1,548 real and 1,524 fake images. 

This dataset is constructed from eight publicly available standard datasets: Celeb-DF-v1\cite{Celeb_DF_cvpr20}, Celeb-DF-v2\cite{Celeb_DF_cvpr20}, FaceForensics++\cite{rössler2019faceforensicslearningdetectmanipulated}, DeepfakeDetection\cite{DeepfakeBench_YAN_NEURIPS2023}, FaceShifter\cite{li2020faceshifterhighfidelityocclusion}, UADFV\cite{DeepfakeBench_YAN_NEURIPS2023}, Deepfake Detection Challenge Preview\cite{dolhansky2019deepfakedetectionchallengedfdc}, and Deepfake Detection Challenge\cite{DeepfakeBench_YAN_NEURIPS2023}. To ensure fair evaluation, the competition organizers pre-processed all datasets consistently to eliminate dependencies on the adopted pre-processing methods. They also anonymized the file names and removed any indicators that could reveal the source dataset.

\subsection{Secondary Dataset Generation}
To ensure our model could effectively generalize to a wider range of forgery techniques, we generated additional fake images by applying several modern deepfake generation methods solely on the real images from the competition dataset. By incorporating fake images generated with more advanced techniques, we aimed to make our model more robust against diverse and evolving deepfake patterns. Specifically, we employed seven deepfake generation methods categorized into three groups: face-swapping, face-reenactment, and face-editing. From the face-swapping category, we selected E4S\cite{liu2023finegrainedfaceswappingregional}, FaceDancer\cite{rosberg2022facedancerposeocclusionawarehigh}, BlendFace\cite{shiohara2023blendfaceredesigningidentityencoders}, and InSwapper\cite{inswapper}. For face-reenactment, we utilized HyperReenact\cite{bounareli2023hyperreenact}, while from the face-editing category, we leveraged e4e\cite{tov2021designingencoderstyleganimage} and StyleCLIP\cite{patashnik2021stylecliptextdrivenmanipulationstylegan}. Each method contributed approximately 2,000 fake images, resulting in an additional 12,200 samples. This augmentation enriched the dataset with diverse forgery techniques, enhancing the model's ability to detect manipulations across various categories.

\subsection{Offline and Online Augmentation}
We employed both offline and online augmentation strategies using \textit{Albumentations} library to enhance the robustness of our model and improve its ability to detect real images.

\textbf{Offline Augmentation:} For real images in the original dataset, we applied offline augmentation to generate an additional 21,335 real examples. Offline augmentation involves applying transformations like random rotation, brightness adjustment, hue shift, and saturation changes to original images and saving the resulting augmented images. These augmented images were later included as real examples during training. This approach significantly increases the diversity of real image samples, helping the model generalize better to unseen data and reducing overfitting.

\textbf{Online Augmentation:} During training, we also implemented online augmentation to introduce variability dynamically. In this strategy, we applied augmentation with a pre-fixed probability for each selected image. If augmentation was triggered, one of six predefined augmentation pipelines was randomly chosen. Each pipeline included a single transformation, such as brightness adjustment, hue shift, saturation change, rotation, or flipping. Online augmentation ensures continuous diversity in training data, making the model more adaptable to real-world variations.

Both offline and online augmentations play a crucial role in ensuring the model’s robustness.

\subsection{Models Considered}
Our study employed several advanced backbone models, each with distinct architectures and capabilities. These models were sourced from the HuggingFace\cite{huggingface} and TIMM\cite{timm} libraries, which provide pretrained weights and efficient implementations for state-of-the-art deep learning models. The models include:

\begin{itemize}
    \item \textbf{MaxViT}: MaxViT\cite{tu2022maxvitmultiaxisvisiontransformer} is a hybrid model that combines convolutional layers and Vision Transformers, designed to balance efficiency and performance. It incorporates a novel multi-axis self-attention mechanism (Max-SA), enabling the model to effectively capture both local patterns and global interactions while reducing computational complexity. MaxViT achieves high performance on image classification tasks by ensuring spatial consistency across feature representations. pretrained on the ImageNet-21K dataset, MaxViT leverages extensive training to extract robust features, making it well-suited for detecting subtle manipulations in deepfake images. By focusing on localized artifacts and maintaining global coherence, MaxViT plays a crucial role in our ensemble. 

    \item \textbf{EVA-02}: EVA-02\cite{Fang_2024} is a next-generation Transformer-based visual representation model pretrained to reconstruct robust language-aligned vision features via masked image modeling. With its updated architecture and extensive pretraining, EVA-02 demonstrates superior generalization capabilities across various vision tasks. It achieves remarkable performance despite requiring fewer parameters and computational resources compared to prior state-of-the-art models. pretrained on the ImageNet-22K dataset, EVA-02 effectively captures global semantic relationships, making it ideal for identifying inconsistencies in facial proportions, lighting, or context in deepfake images. Its ability to extract robust global features complements the localized focus of 
    MaxViT, ensuring a balanced feature representation in our ensemble. 

    \item \textbf{CoAtNet}: CoAtNet \cite{dai2021coatnetmarryingconvolutionattention} is a hybrid model designed to unify the strengths of convolutional networks and Transformers through depthwise convolution and self-attention. CoAtNet introduces a structured inductive bias that enhances generalization, capacity, and efficiency by vertically stacking convolution and attention layers. pretrained on the ImageNet-12K dataset, CoAtNet achieves state-of-the-art performance with fewer resource requirements. Its ability to effectively capture features at various scales makes it particularly suitable for addressing both local and global manipulations present in deepfake images. CoAtNet bridges the gap between convolutional networks and Transformers, making it an essential component of our ensemble. 
\end{itemize}

The actual versions of the models, the datasets they were pretrained on, and their Top-1 and Top-5 accuracies on the ImageNet\cite{deng2009imagenet} dataset have been provided in Table~\ref{tab:model_comparison}. Notably, in our adaptation, all input images were resized to \(256 \times 256\) pixels before being passed to the models.

\begin{table*}[h!]
\centering
\renewcommand{\arraystretch}{1.4}
\setlength{\tabcolsep}{10pt}
\caption{Model comparison by pretrained dataset, trainable parameters, and performance metrics. Top-1 and Top-5 accuracy values are reported on the ImageNet dataset \cite{deng2009imagenet}}
\label{tab:model_comparison}
\begin{tabular}{p{3.8cm}|l|c|c|c}
\hline
\textbf{Model Name} & \textbf{Pretrained On} & \textbf{Trainable Parameters (M)} & \textbf{Top-1 Accuracy} & \textbf{Top-5 Accuracy} \\
\hline
EVA-02 Base Patch14 448 & ImageNet-22k & 87.1 & 0.887 & 0.987 \\
\hline
CoAtNet RMLP-2 RW 384  & ImageNet-12k & 73.9 & 0.8739 & 0.9831 \\
\hline
MaxViT Base TF 512      & ImageNet-21k & 119.9 & 0.8820 & 0.9853 \\
\hline
\end{tabular}
\end{table*}

\begin{table*}[h!]
\centering
\renewcommand{\arraystretch}{1.2}
\setlength{\tabcolsep}{10pt}
\caption{Hyperparameters for Backbone Models and Classifiers}
\label{tab:hyperparameters}
\begin{tabular}{l|c|c|c|c|c|c|c}
\hline
\multirow{2}{*}{\textbf{Model}} & \multicolumn{4}{c|}{\textbf{Backbone}} & \multicolumn{3}{c}{\textbf{Classifier}} \\ \cline{2-8} 
                                & \textbf{Batch Size} & \textbf{Epochs} & \textbf{Learning Rate} & \textbf{Weight Decay} & \textbf{Batch Size} & \textbf{Epochs} & \textbf{Learning Rate} \\ \hline
MaxViT                          & 16                  & 4               & 3e-5                   & 1e-2                  & 16                  & 8               & 5e-5                   \\ \hline
CoAtNet                         & 16                  & 2               & 3e-5                   & 1e-2                  & 16                  & 8              & 5e-5                   \\ \hline
EVA-02                          & 16                  & 6               & 3e-5                   & 1e-2                  & 16                  & 7               & 5e-5                   \\ \hline
\end{tabular}
\end{table*}

\subsection{Hyperparameters}

For training, we utilized the AdamW optimizer from the \textit{PyTorch} library for the backbone and the Adam optimizer for the classifier. Additionally, we employed a ReduceLROnPlateau scheduler to dynamically adjust the learning rate by a factor of 0.5 when the validation loss plateaus, ensuring efficient convergence. The scheduler is configured with a patience of 1 epoch. The hyperparameters for the backbone models and the task-specific classifiers are summarized in Table~\ref{tab:hyperparameters}.

\subsection{Supervised Contrastive Loss}
Supervised contrastive (SupCon) loss\cite{khosla2021supervisedcontrastivelearning}\cite{supcontrast} is used to create distinct clusters for real and fake images in the embedding space. By pulling together embeddings of images with the same label (real or fake) and pushing apart embeddings of different labels, it ensures that real and fake images are well-separated. In our implementation, we first normalize the embeddings to maintain consistency. A similarity matrix is then computed to evaluate pairwise relationships among all samples in a batch. Using the labels, a mask is created to identify positive pairs (images with the same label) and negative pairs (images with different labels). The loss function maximizes similarity for positive pairs and minimizes it for negative pairs. This structured separation in the embedding space serves as a strong foundation for the classifier in the next stage, enabling it to distinguish between real and fake images effectively.






These formulations allow for generalization to arbitrary numbers of positives and negatives, ensuring a robust clustering of representation space. They retain the intrinsic ability to perform hard positive/negative mining, enabling the model to focus on difficult examples without explicit mining strategies.

\subsection{Training the Backbone}
In the first stage of training (illustrated in Figure~\ref{fig:stage1}), we utilize backbone models such as MaxViT, EVA-02, and CoAtNet, fine-tuned using supervised contrastive loss. We employ a sampling strategy to address the class imbalance in the dataset instead of using the full dataset. Specifically, for each backbone, we include all real images (42,690 original and 21,335 augmented, totaling 64,025) while sampling the fake images into three distinct subsets, each containing 73,157 fake images. Additionally, we incorporate 12,200 generated fake images from the provided dataset. As a result, each backbone is trained on 64,025 real images and 85,357 fake images, with different backbones trained on distinct subsets of fake images.

This sampling strategy mitigates the significant class imbalance between real and fake images, which could otherwise lead to biased learning. During training, we extract feature embeddings directly from the backbone models and use those to compute the supervised contrastive loss, which ensures that embeddings of real images cluster tightly together while maintaining clear separation from those of fake images. This approach enhances the discriminative quality of the feature embeddings, making them highly effective for downstream tasks.

The t-SNE visualizations, presented in Figure~\ref{fig:tsne_visualization}, further validate the effectiveness of this training approach. Before training, the embeddings of real and fake images significantly overlap, as seen in Figures~\ref{fig:tsne_visualization}a, \ref{fig:tsne_visualization}b, and \ref{fig:tsne_visualization}c. This overlap indicates the lack of meaningful separation in the feature space. After training, as shown in Figures~\ref{fig:tsne_visualization}d, \ref{fig:tsne_visualization}e, and \ref{fig:tsne_visualization}f, the embeddings exhibit distinct clustering for real and fake images. The supervised contrastive loss enables the backbone models to learn robust and discriminative features, which result in well-separated clusters, laying a strong foundation for accurate classification in subsequent stages.

\subsection{Training the Classifier}
In the second stage of training (depicted in Figure~\ref{fig:stage2}), we utilize the pretrained backbone models, including MaxViT, EVA-02, and CoAtNet, with their parameters frozen. An MLP (Multi-Layer Perceptron) with ReLU activation, Batch Normalization, and Dropout layers is added on top of the frozen backbone to serve as the classifier. For the loss function, we use the BCEWithLogitsLoss function with Adam optimizer.

Feature embeddings are generated by passing images through the frozen backbone. These embeddings are then used as input to the classifier, which is trained to distinguish between real and fake images. This approach ensures that the classifier utilizes the well-separated embeddings produced by the backbone models, leading to accurate classification.

\subsection{Ensembling the Models}
To enhance the reliability of our deepfake detection system (as shown in Figure~\ref{fig:stage3}), we ensemble three backbone models: EVA-02, MaxViT, and CoAtNet. These models were chosen for their complementary strengths— EVA-02 specializes in capturing global relationships, MaxViT balances local and global feature extraction, and CoAtNet effectively handles multi-scale features with its hybrid architecture. Together, they provide a robust framework for detecting subtle manipulations in deepfake images.

For ensembling, we use a Majority Voting strategy to decide whether the input image is real or fake. Once a decision is made, the probability of being a real image is rendered as follows. If the image has been decided as a real image (i.e., two out of three models gave probability $> 0.5$), then the highest probability output among the models is returned. If, on the other hand, the image has been determined to be fake, then the lowest probability output among the models is returned. This strategy effectively combines the strengths of the individual models, improving the system’s overall performance and adaptability in diverse scenarios.

\subsection{Hardware Specification and Environment Setup}
For training, we utilized Kaggle NVIDIA T4x2 GPU instances, which provided sufficient computational resources, including GPU, RAM, and disk space\cite{kaggle}.

The training times for the backbone models, including pretraining (training the backbone) and training (training the classifier), are summarized in Table~\ref{tab:training_times}. The table highlights the efficiency and computational requirements of each model, with CoAtNet requiring the least time for both pretraining and training stages, while MaxViT consumed the most time due to its higher computational complexity. The time required to process one test file is 0.045 sec.

\begin{table}[h!]
\centering
\renewcommand{\arraystretch}{1.2}
\setlength{\tabcolsep}{10pt}
\caption{Per-epoch training times for backbone models. Pretraining refers to training the backbone, while training refers to training the classifier.}
\label{tab:training_times}
\begin{tabular}{l|c|c}
\hline
\textbf{Model} & \textbf{Pretraining (per epoch)} & \textbf{Training (per epoch)} \\
\hline
CoAtNet  & 1 hour 35 minutes & 59 minutes \\
\hline
EVA-02   & 1 hour 45 minutes & 1 hour \\
\hline
MaxViT   & 2 hours 45 minutes & 1 hour 30 minutes \\
\hline
\end{tabular}
\end{table}

\section{Results and Discussions}

\subsection{Baseline Comparison}
To validate the effectiveness of our proposed method, we performed a baseline comparison against widely-used backbone models, including ResNet50, ResNet152, ResNet101, InceptionV3, and InceptionResNetV2, as shown in Table~\ref{tab:baseline_comparison}. The comparison metrics include the Top-1 Accuracy on the ImageNet dataset and accuracy on the competition validation dataset (DFWild). These results provide a benchmark for assessing the robustness and generalization capability of different models in deepfake detection tasks.

Among the baseline models, InceptionResNetV2 demonstrated the highest accuracy on the DFWild dataset with a score of \(0.8724\), followed closely by ResNet50 at \(0.8721\). This indicates the strong potential of these models in capturing both local and global feature representations. ResNet152 and ResNet101 showed slightly lower accuracies of \(0.8574\) and \(0.8369\), respectively. InceptionV3 achieved an accuracy of \(0.8698\), highlighting its balanced trade-off between parameter efficiency and performance.

Our proposed method significantly outperformed all baseline models by achieving an accuracy of \(0.9583\), as detailed in Table~\ref{tab:ensembling_results}. This remarkable improvement can be attributed to the ensemble of advanced backbone models (MaxViT, CoAtNet, and EVA-02), combined with supervised contrastive loss and comprehensive augmentation strategies. These enhancements enable our framework to effectively generalize across diverse datasets and address the challenges of deepfake detection with superior accuracy and robustness.

\begin{table}[h!]
\centering
\renewcommand{\arraystretch}{1.2}
\setlength{\tabcolsep}{4pt}
\caption{Baseline Comparison of Backbone Models. Accuracy on ImageNet and DFWild datasets is shown.}
\label{tab:baseline_comparison}
\begin{tabular}{l|c|c|c}
\hline
\multirow{2}{*}{\textbf{Model Name}} & \multirow{2}{*}{\textbf{Parameters (M)}} & \textbf{ImageNet Top-1} & \textbf{DFWild} \\ 
                                     &                                          & \textbf{Accuracy}       & \textbf{Accuracy} \\ 
\hline
ResNet50            & 25.57                  & 0.7530                  & 0.8721          \\
ResNet152           & 66.84                  & 0.7857                  & 0.8574          \\
ResNet101           & 44.57                  & 0.7825                  & 0.8369          \\
InceptionV3         & 23.83                  & 0.7895                  & 0.8698          \\
InceptionResNetV2   & 55.84                  & \textbf{0.8046}                  & \textbf{0.8724}          \\
\hline
\end{tabular}
\end{table}

\subsection{Model Selection Results}
We evaluated advanced backbone architectures, including MaxViT, CoAtNet, EVA-02, ConvNeXtV2, and EfficientNetV2, based on accuracy, AUC, and parameter efficiency, as summarized in Table~\ref{tab:model_selection_results}.

MaxViT achieved the highest accuracy (\(0.9508\)) with its multi-axis attention mechanism excelling in feature extraction. CoAtNet followed with \(0.9447\) accuracy and \(0.9787\) AUC, leveraging its hybrid convolution-attention design for multi-scale representation. EVA-02 achieved \(0.9248\) accuracy and the highest AUC (\(0.9791\)), benefiting from robust pretraining with masked image modeling.

ConvNeXtV2 and EfficientNetV2 displayed competitive performance but lagged behind the top three models in terms of accuracy and AUC. These findings confirmed the choice of MaxViT, CoAtNet, and EVA-02 as the backbone models for our proposed framework due to their complementary strengths and superior metrics.

\begin{table}[h!]
\centering
\renewcommand{\arraystretch}{1.2}
\setlength{\tabcolsep}{8pt}
\caption{Model Selection Results (Validation set). Accuracy and AUC values are reported.}
\label{tab:model_selection_results}
\begin{tabular}{l|c|c|c}
\hline
\textbf{Model}         & \textbf{Parameters (M)} & \textbf{Accuracy}  & \textbf{AUC} \\
\hline
MaxViT                 & 119.9                   & \(0.9508\)         & \(0.9762\)   \\
CoAtNet                & 73.9                    & \(0.9447\)         & \(0.9787\)   \\
EVA-02                 & 87.1                    & \(0.9248\)         & \(0.9791\)   \\
EfficientNetV2         & 21.6                    & \(0.8779\)         & \(0.9686\)   \\
ConvNeXtV2             & 28.6                    & \(0.9118\)         & \(0.9670\)   \\
\hline
\end{tabular}
\end{table}

\begin{figure*}[h!]
    \centering
    \begin{tabular}{ccc}
        \includegraphics[width=0.3\textwidth]{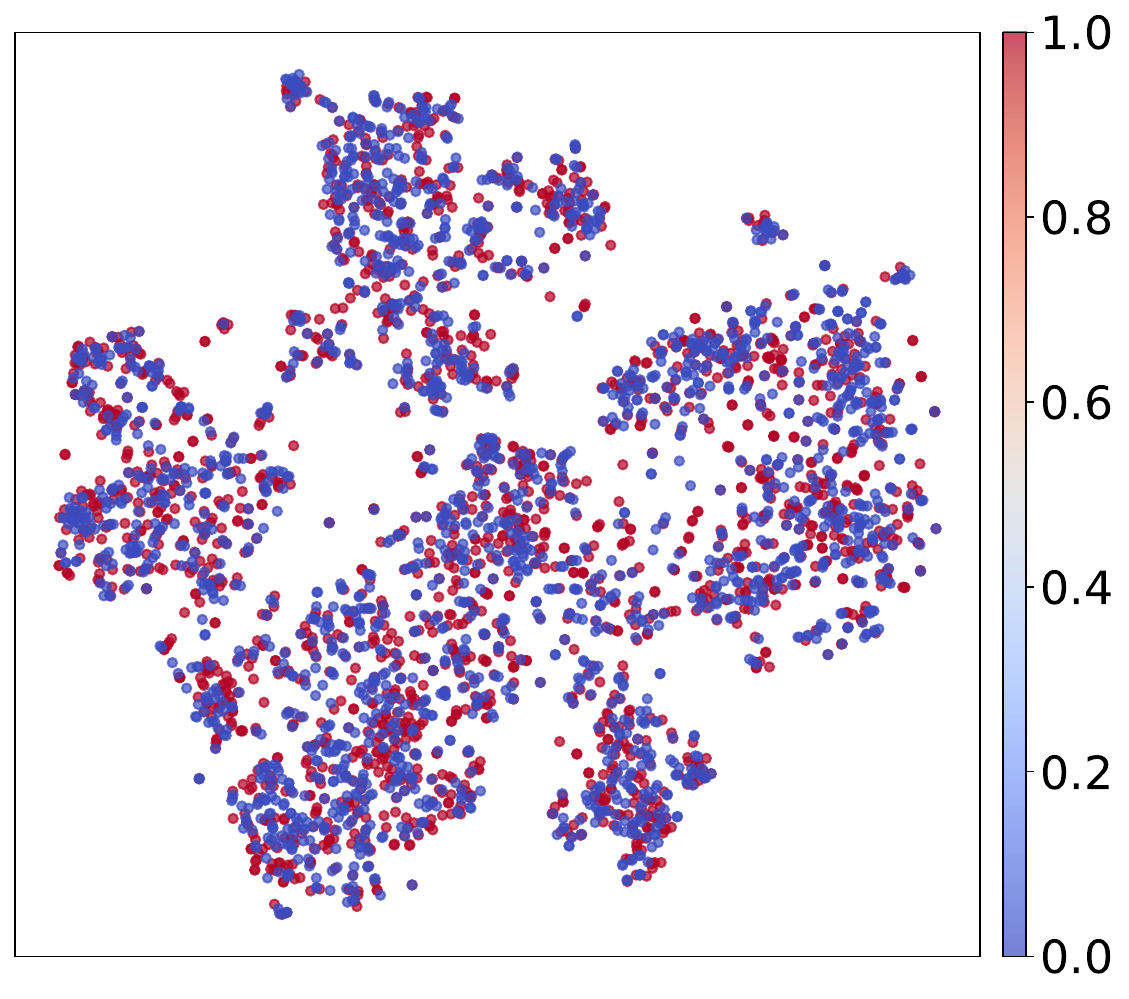} &
        \includegraphics[width=0.3\textwidth]{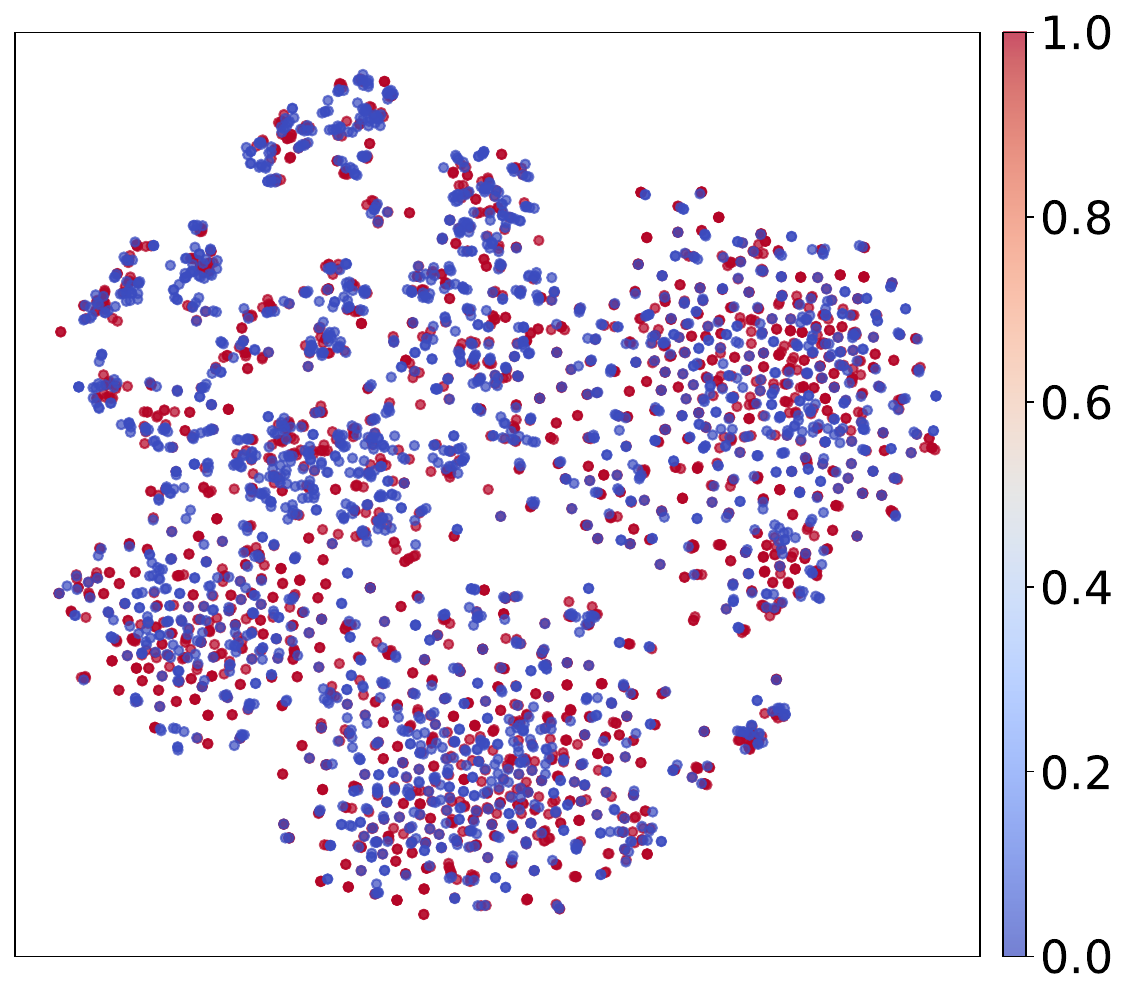} &
        \includegraphics[width=0.3\textwidth]{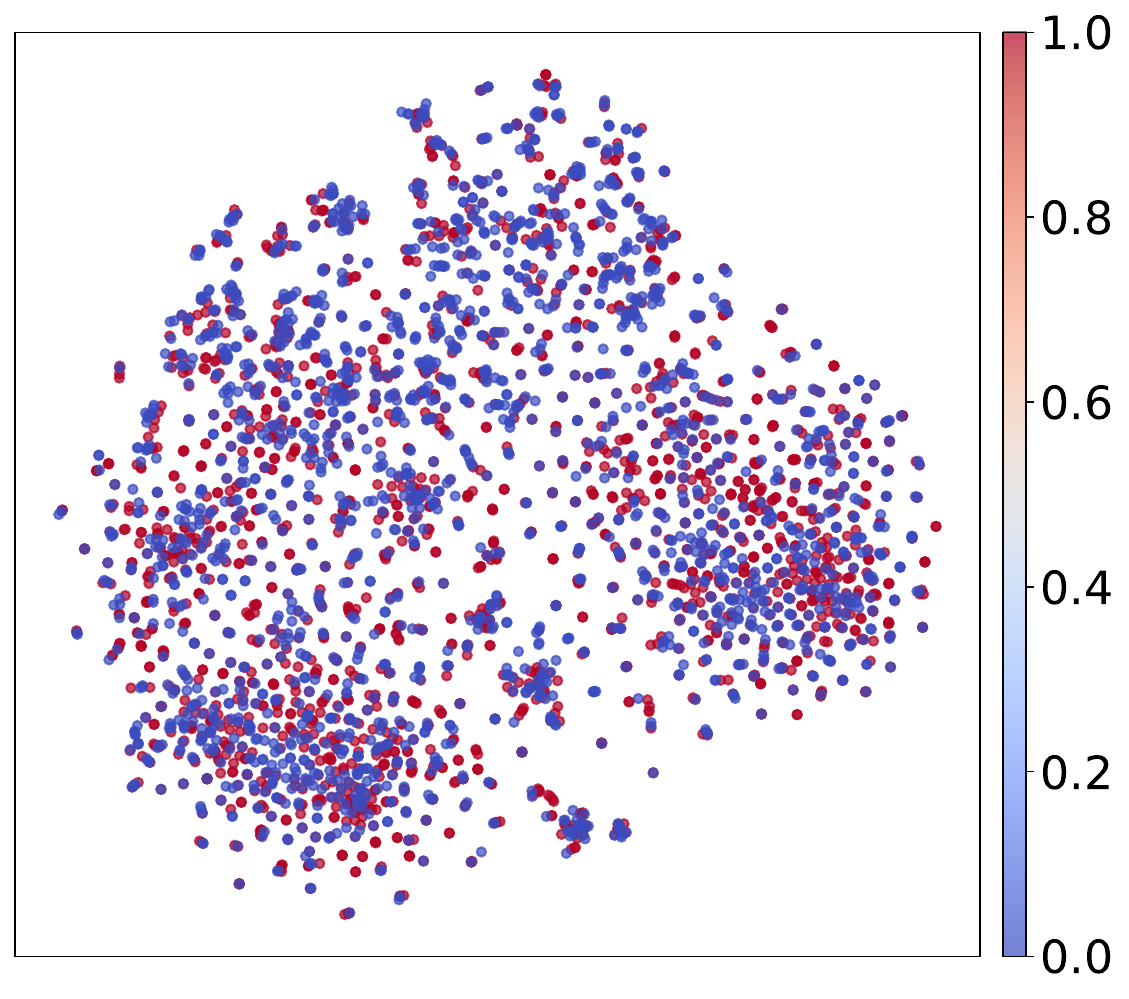} \\
        (a) MaxViT (Before Training) & (b) EVA-02 (Before Training) & (c) CoAtNet (Before Training) \\
        \includegraphics[width=0.3\textwidth]{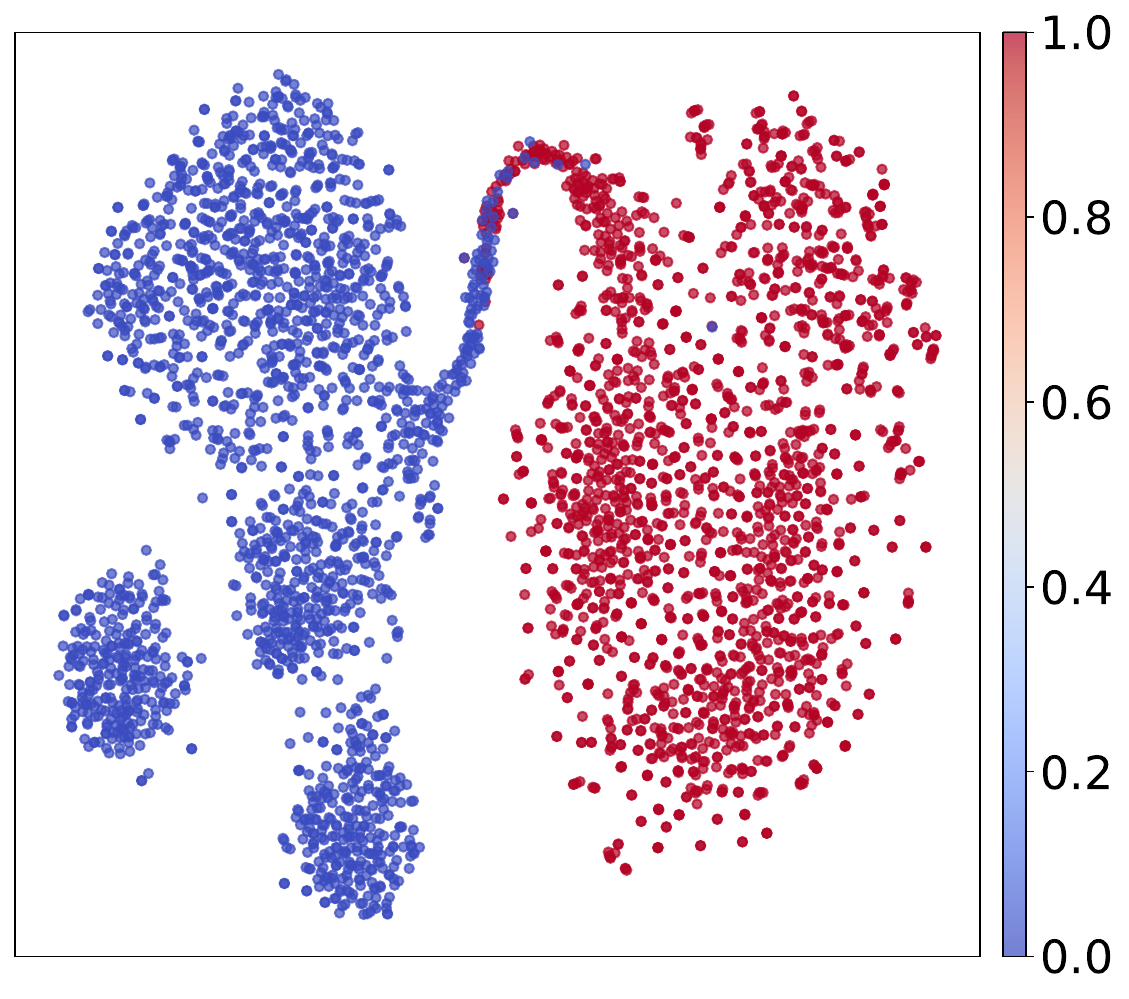} &
        \includegraphics[width=0.3\textwidth]{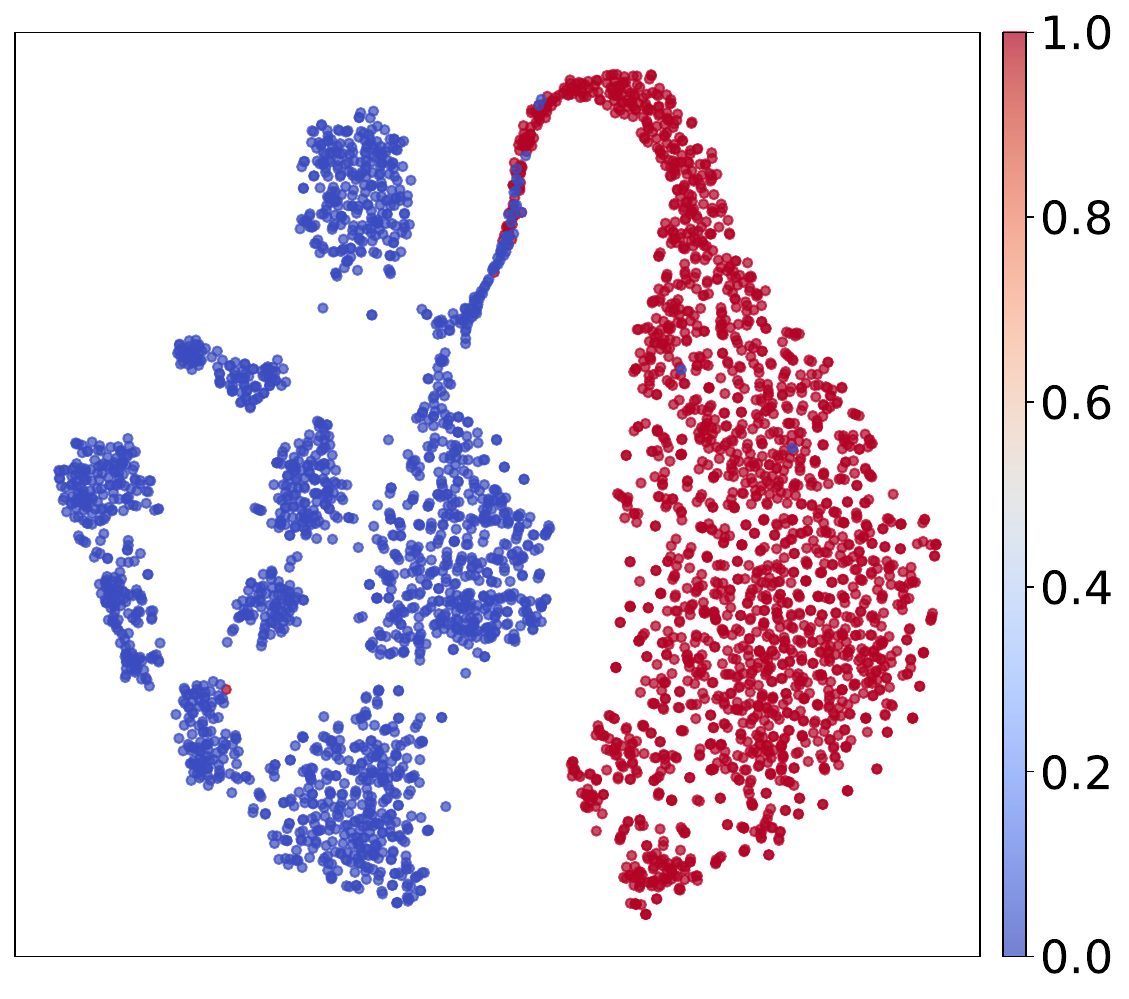} &
        \includegraphics[width=0.3\textwidth]{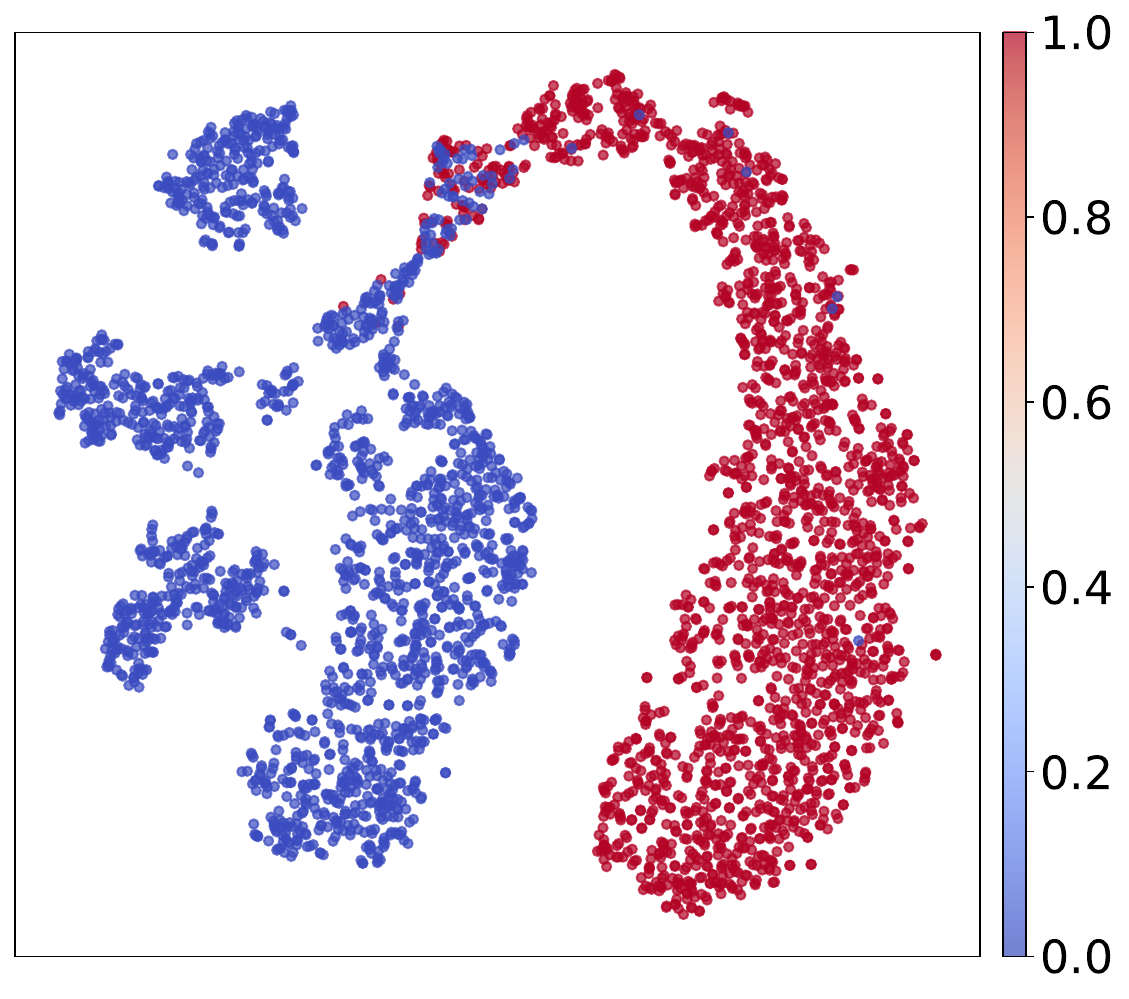} \\
        (d) MaxViT (After Training) & (e) EVA-02 (After Training) & (f) CoAtNet (After Training) \\
    \end{tabular}
    \caption{t-SNE visualization of feature embeddings before and after training for MaxViT, EVA-02, and CoAtNet. The plots illustrate how the embeddings of real (Label 1 (red)) and fake (Label 0 (blue)) images become more separable after training. To generate these visualizations, we randomly selected 2,000 real and 2,000 fake images from the training dataset.}
    \label{fig:tsne_visualization}
\end{figure*}

\subsection{Ensembling Results}
To further enhance performance, we combined the predictions of MaxViT, CoAtNet, and EVA-02 using an ensemble strategy. This approach utilized majority voting of the models’ output probabilities, capitalizing on their complementary strengths. 

As shown in Table~\ref{tab:ensembling_results}, the ensemble outperformed individual models, achieving an F1 score of 0.9586, an accuracy of 0.9583, and an AUC of 0.9807. The ensemble demonstrated significant improvements in generalization across diverse scenarios, confirming the efficacy of combining models with distinct feature-capturing abilities. This highlights the importance of leveraging both local and global feature representations for robust deepfake detection.

\begin{table}[h!]
\centering
\renewcommand{\arraystretch}{1.2}
\setlength{\tabcolsep}{6pt}
\caption{Ensembling Results (Validation set)}
\label{tab:ensembling_results}
\begin{tabular}{l|c|c|c|c|c}
\hline
\textbf{Model} & \textbf{Accuracy} & \textbf{F1 Score} & \textbf{Precision} & \textbf{Recall} & \textbf{AUC} \\
\hline
MaxViT         & 0.9508                  & 0.9507            & 0.9604             & 0.9412          & 0.9762       \\
CoAtNet        & 0.9447                  & 0.9453            & 0.9422             & 0.9483          & 0.9787       \\
EVA-02         & 0.9248                  & 0.9222            & \textbf{0.9628}             & 0.8850          & 0.9791       \\
\textbf{Ensemble} & \textbf{0.9583}       & \textbf{0.9586}   & 0.9604   & \textbf{0.9567} & \textbf{0.9807} \\
\hline
\end{tabular}
\end{table}

The t-SNE visualizations (Figure~\ref{fig:tsne_visualization}) provide valuable insights into the effectiveness of our models. For real images, a single cohesive cluster is observed, reflecting the uniformity in the characteristics of real images. In contrast, fake images form multiple distinct clusters in the embedding space, likely corresponding to the various deepfake generation methods used. This suggests that each deepfake generation technique introduces unique artifacts that the backbone models can detect and represent differently.

After training, all three backbone models—MaxViT, CoAtNet, and EVA-02—produced well-separated embeddings for real and fake images, as seen in Figure~\ref{fig:tsne_visualization}d, \ref{fig:tsne_visualization}e, and \ref{fig:tsne_visualization}f. However, the ensemble strategy further leverages these representations, combining the strengths of the models to produce more consistent and robust separation. By capturing complementary features—local patterns from MaxViT, multi-scale representations from CoAtNet, and global contextual understanding from EVA-02—the ensemble excels in generalization and reliability across diverse datasets.

\subsection{Performance on Diverse Scenarios}

To ensure the robustness of our model across diverse scenarios, such as variations in age and ethnicity, we employed both offline and online augmentations, incorporating transformations like brightness adjustments, hue-saturation shifts, and rotations to simulate real-world diversity.

Additionally, we used regularization techniques such as AdamW's weight decay, BatchNorm, and dropout to prevent overfitting and enhance generalization. Weight decay penalizes large weight values, promoting simpler and more robust representations. BatchNorm stabilizes training by normalizing activations, while dropout deactivates random neurons, encouraging the model to learn generalized patterns rather than specific features. We expect that these strategies will make our model both generalizable and fair across diverse scenarios.

The ensembling of backbone models—MaxViT, CoAtNet, and EVA-02—played a pivotal role in capturing diverse feature representations. MaxViT effectively balanced local and global feature extraction with its multi-axis attention mechanism, making it adept at identifying fine-grained manipulations. The hybrid architecture of depthwise convolution and attention layers of CoAtNet ensured efficient multi-scale feature representation. EVA-02, leveraging its pretraining on large-scale datasets, provided superior global contextual understanding. This complementary combination resulted in consistent performance improvements across all scenarios, validating the effectiveness of the proposed ensemble strategy.

\subsection{Ablation Study}
The ablation study, shown in Table~\ref{tab:ablation_study}, evaluates the impact of key components in our method using CoAtNet as the backbone. CoAtNet was chosen due to its efficiency in terms of training time and resource usage. This study demonstrates how each component of our pipeline contributes to the overall performance.

Removing offline or online augmentations resulted in significant accuracy drops, with \(0.9303\) and \(0.8659\), respectively, emphasizing the critical role of data diversity in enhancing generalization. Replacing SupCon loss with BCE loss led to a reduced accuracy of \(0.9163\), showcasing the advantage of SupCon loss in improving embedding separability. The proposed method on CoAtNet, combining SupCon loss and augmentations, achieved an accuracy of \(0.9447\), validating the effectiveness of these components working together.

\begin{table}[h!]
\centering
\renewcommand{\arraystretch}{1.2}
\setlength{\tabcolsep}{6pt}
\caption{Ablation study of the proposed method with CoAtNet.}
\label{tab:ablation_study}
\begin{tabular}{l|c|c|c}
\hline
\textbf{Configuration} & \textbf{Accuracy} & \textbf{F1 Score} & \textbf{Precision}\\
\hline
No Offline Augmentation & 0.9303 & 0.9339 & 0.8947 \\
No Online Augmentation  & 0.8659 & 0.8574 & 0.9232 \\
SupCon Loss replaced by BCE loss  & 0.9163 & 0.9131 & \textbf{0.9581} \\
\textbf{Ours (SupCon + Augmentations)} & \textbf{0.9447} & \textbf{0.9453} & 0.9422 \\
\hline
\end{tabular}
\end{table}

\subsection{Insights and Observations}
The experimental results provide several key insights:
\begin{itemize}
    \item The use of supervised contrastive loss enabled clear separation between real and fake embeddings, improving the classifier’s ability to differentiate effectively.
    \item Augmentation techniques, coupled with balanced sampling, mitigated dataset biases and enhanced the model’s performance on unseen data.
    \item Ensembling backbone models with distinct strengths minimized biases of individual models, improving overall accuracy.
    \item Utilizing pretrained models from Hugging Face\cite{huggingface} and TIMM\cite{timm} libraries accelerated training and ensured high performance with reduced computational requirements.
\end{itemize}

These observations underscore the effectiveness of our approach in tackling the challenges of deepfake detection, demonstrating its applicability to real-world scenarios and setting a new benchmark for this task.

\section{Conclusion}
In this report, we presented our approach for deepfake detection in the IEEE SP Cup 2025: Deepfake Face Detection in the Wild (DFWild-Cup). Our method integrates three advanced models—MaxViT, EVA-02, and CoAtNet—capitalizing on their strengths in capturing both local and global features. Combining these models through ensembling has improved accuracy, ensuring better performance in diverse and unseen datasets. The supervised contrastive loss was utilized to create well-separated embedding spaces, which significantly enhanced the classification process. Additionally, both offline and online data augmentation strategies were applied to improve the system’s adaptability to real-world conditions. The methodology emphasizes an efficient combination of powerful models and strategic training techniques, achieving strong results in deepfake detection. This approach can serve as a foundation for further advancements in image classification tasks requiring high precision and adaptability.

\section*{Acknowledgment}

We want to thank the organizers of SP Cup 2025\cite{ieeespcup} as well as 
IEEE for launching a competition with a problem that is 
really important in the context of our time. 

\bibliographystyle{IEEEtran}
\bibliography{references}

\end{document}